\documentclass{article}
\usepackage[preprint]{neurips_2026}

\usepackage{tabularx}
\usepackage[table,xcdraw,dvipsnames]{xcolor} 
\usepackage[utf8]{inputenc} 
\usepackage[T1]{fontenc}
\usepackage{graphicx}
\usepackage{glossaries}
\usepackage{soul} 
\usepackage[bookmarksdepth=2]{hyperref} 
\usepackage{bbding}
\usepackage{float} 
\usepackage[many]{tcolorbox} 
\usepackage{enumitem} 
\usepackage{ifthen} 
\usepackage{amssymb} 
\usepackage{booktabs}
\usepackage{graphicx}
\usepackage{makecell}
\usepackage{amsmath}
\usepackage{url}
\usepackage[export]{adjustbox}
\usepackage{ragged2e} 
\usepackage{relsize} 
\usepackage{multirow} 
\usepackage{pbox} 
\usepackage{balance} 
\usepackage{moresize} 
\usepackage{xspace}

\usepackage{placeins} 

\newtcolorbox{mybox}[2][]{
top=0.15in,left=4pt,right=4pt,bottom=4pt,
fonttitle=\bfseries,
colbacktitle=gray,
colback=gray!5,
colframe=gray!40!black,
enhanced,
attach boxed title to top left={xshift=1.5em,yshift=-\tcboxedtitleheight/2},
boxed title style={size=small},
drop shadow={black!50!white},
title=#2,#1}

\newcommand{\countobservations}{
    \def \countobservations{1}
}
\newcounter{observation}
\countobservations

\newcommand{\countimplications}{
    \def \countimplications{1}
}
\newcounter{implication}
\countimplications

\newboolean{showcomments}
\setboolean{showcomments}{true}

\ifthenelse{\boolean{showcomments}}
{\newcommand{\nbc}[3]{
 {\colorbox{#3}{\bfseries\sffamily\scriptsize\textcolor{white}{#1}}}
 {\textcolor{#3}{\sf\small$\blacktriangleright$\textit{#2}$\blacktriangleleft$}}
 }
}
{\newcommand{\nbc}[3]{}
 }

\newcommand\ahsan[1]{\nbc{Ahsan}{#1}{blue}}

\newcommand\nakhla[1]{\nbc{Nakhla}{#1}{teal}}
\newcommand{\zhijie}[1]{\nbc{Zhijie}{#1}{olive}}

\newcommand{\tool}{\textit{FALAT}\xspace}

\newcommand{\smalltt}[1]{\ifmmode{\mbox{\smaller\texttt{#1}}}\else{\smaller\tt #1}\fi}

\newcommand{\phead}[1]{\noindent {\bf #1}}
\newcolumntype{L}[1]{>{\raggedright\let\newline\\\arraybackslash\hspace{0pt}}m{#1}}
\newcolumntype{C}[1]{>{\centering\let\newline\\\arraybackslash\hspace{0pt}}m{#1}}
\newcolumntype{R}[1]{>{\raggedleft\let\newline\\\arraybackslash\hspace{0pt}}m{#1}}

\DeclareMathOperator*{\argmin}{arg\,min}

\title{{\tool}: Tracing Failures in LLM Agent Trajectories via Dependency-Guided Search}

\author{
Md Nakhla Rafi \\
SPEAR Lab\\
  Concordia University\\ Montreal, Canada\\
  \texttt{mdnakhla.rafi@mail.concordia.ca} \\
\And
Md Ahasanuzzaman \\
SPEAR Lab\\
Concordia University\\ Montreal, Canada\\
\texttt{m\_ahasa@live.concordia.ca} \\
\AND
Dong Jae Kim \\
DePaul University \\
Chicago, USA \\
\texttt{dkim121@depaul.edu}
\And
Zhijie Wang \\
Concordia University\\ Montreal, Canada\\
\texttt{zhijie.wang@concordia.ca}
\AND
Tse-Hsun Chen \\
SPEAR Lab\\
Concordia University\\ Montreal, Canada\\
\texttt{peterc@encs.concordia.ca
}
}

\begin{document}

\maketitle
\begin{abstract}
LLM-based agents increasingly solve complex tasks through long trajectories involving reasoning steps, tool calls, and inter-agent communication. However, when these agents fail, it is often unclear which agent caused the failure and which step introduced the decisive error. This attribution problem is challenging because mistakes can propagate across the trajectory: later actions may appear incorrect, but only because they depend on an earlier corrupted state. Therefore, failure attribution cannot be treated as independent step-level classification. We propose \tool, a diagnostic framework for failure attribution in LLM agent trajectories. \tool frames attribution as a dependency-guided search problem. It first constructs an expectation of how the task should be solved and uses this expectation to identify suspicious regions in the trajectory. It then traces dependencies among decisions, tool outputs, and agent messages to distinguish error-introducing steps from steps that merely inherit or propagate prior mistakes. Finally, \tool evaluates whether correcting a candidate step would be sufficient to recover the expected outcome, allowing it to identify both the responsible agent and the decisive failure step. We evaluate \tool on the Who\&When benchmark, which includes both algorithm-generated and hand-crafted multi-agent failure trajectories. The results show that \tool consistently improves responsible-agent and decisive-step attribution. Its best configurations achieve 46.0\% step-level accuracy on algorithm-generated trajectories and 29.1\% on the more challenging hand-crafted trajectories, outperforming specialized attribution baselines and direct prompting with standalone LLMs. These findings suggest that dependency-aware reasoning is essential for reliable failure diagnosis in LLM agent systems.
\end{abstract}

\section{Introduction}
\label{sec:introduction}

Large language models (LLMs) have enabled a new class of agentic systems that solve complex tasks through iterative reasoning, tool use, and environment interaction. Modern software engineering agents such as OpenAI Codex and Claude Code can autonomously edit code, invoke tools and scripts, and execute tests over long execution horizons. Such a sequence of actions generated and executed by these agents is also referred to as a \emph{trajectory}.

Despite rapid progress, state-of-the-art (SOTA) agents still fall short on many real-world tasks, particularly those involving long-range dependencies, under-specified requirements, and multi-step execution chains~\citep{jimenez2023swe, yu2025utboost}. 
Understanding \emph{why} an agent failed is thus increasingly important as these systems become more autonomous. When a trajectory produces an incorrect output, developers must determine which agent (\emph{who}) introduced the decisive error at which step(s) (\emph{when}), where a decisive error refers to an error whose correction would recover the final outcome of the agentic system.

Addressing this problem, which we refer to as \emph{failure attribution}, is essential for ensuring the reliability of agentic systems. However, decisive error localization is challenging because trajectories are often long, interdependent, and difficult to inspect directly. Moreover, once an early mistake corrupts the execution state, many downstream steps may appear locally reasonable while still propagating the original fault. As a result, the true error source can be obscured by later symptoms.

Recent failure-attribution methods address these issues only partially. Single-pass LLM judges treat trajectories as flattened inputs, making them vulnerable to positional bias~\citep{liu2024lost}, order sensitivity in the presented evidence~\citep{rafi2024order}, and limited hypothesis revision across long chains~\citep{zhu2025llm}. Learning-based tracers~\citep{zhang2025agentracer} independently evaluate candidate steps and predict which one is responsible for the failure. While intuitive, this formulation generalizes poorly in practice~\citep{zainullina2025guided}. 



In this paper, we reformulate failure attribution as \emph{hierarchical dependency-guided search}. Rather than classifying steps independently, \tool performs a top-down search across layers of abstraction and uses typed dependencies to guide error candidate pruning and verification. \tool first constructs a task-conditioned prior and a hierarchical representation of the trajectory to support coarse-to-fine localization. The prior is constructed independently of the trajectory's intermediate steps so that candidate selection is not guided solely by the agent's own flawed reasoning. \tool then constructs typed dependencies among candidate steps to separate possible error sources from downstream carriers and prune candidates whose effects cannot reach the final output. The remaining candidates are ranked and verified by checking whether fixing a candidate would recover the expected outcome. Finally, \tool performs local verification around the predicted decisive step to reduce premature commitment to a plausible but non-decisive candidate.


We evaluate \tool on the Who\&When benchmark~\citep{zhang2025which} using three LLM backbones and compare it against two state-of-the-art attribution baselines and multiple standalone LLM baselines. \tool consistently achieves higher responsible-agent and decisive-step attribution accuracy. On algorithm-generated trajectories, \tool reaches 46.0 step-level accuracy, compared with 37.3 for CHIEF~\citep{wang2026flat} under the same MiniMax 2.5~\citep{minimax2025m25} backbone. On the more challenging hand-crafted trajectories, \tool reaches 29.1 step-level accuracy, compared with 17.0 for CHIEF under DeepSeek V3.2~\citep{liu2025deepseek}.

Our contributions are summarized as follows:
\begin{itemize}

    \item We formulate failure attribution in LLM agent trajectories as
    \emph{hierarchical dependency-guided search}, shifting the task from flat
    step classification to structured diagnosis over candidate steps and their
    dependencies.

    \item We propose {\tool}, a diagnostic framework that uses expected task
    behavior and step dependencies to identify where an error is introduced.

    \item We evaluate {\tool} on the Who\&When benchmark across multiple LLM
    backbones, showing consistent gains over specialized attribution baselines
    and direct prompting with standalone LLMs, especially on the stricter
    step-level attribution task.

\end{itemize}

\section{Hierarchical Dependency-Guided Search for Agent Trajectory Diagnostics}
\label{sec:approach}

In this paper, we reformulate failure attribution as \emph{hierarchical dependency-guided search}. The task is to identify the earliest step, or set of steps, whose correction would change the system's final output from the observed incorrect output $\hat{o}$ to the expected output $o^*$. The search proceeds top-down through layers of abstraction and uses typed dependency relations to prune candidates.


This reformulation addresses two properties of the problem that step-level classification ignores. First, directly searching on trajectories spanning hundreds of steps can be both inefficient and ineffective compared to coarse-to-fine narrowing, 
as shown in classical abstraction hierarchies for planning~\citep{sacerdoti1974planning} and reinforcement learning~\citep{sutton1999between}. 
Second, since errors may propagate across steps, distinguishing origination from propagation requires \emph{deductive} reasoning over dependency structure rather than ranking individual steps independently.



These observations motivate three design principles for \tool.

\begin{itemize}
    \item \textbf{Expectation-guided search.} Candidate error steps should be identified using an external prior over expected behavior. Signals derived solely from the trajectory can be misleading because they may reflect the same flawed assumptions that led the agent to fail.

    \item \textbf{Source-propagation separation.}
    Dependency relations between candidate steps should be typed to distinguish error origination from propagation. Without this distinction, straightforward LLM judgments often mistake downstream propagated errors for decisive causes~\citep{guo2026agent}. 

    \item \textbf{Counterfactual sufficiency.} A candidate is decisive only if fixing it would recover the expected output $o^*$. This counterfactual test holds earlier steps unchanged while allowing later steps to adjust, filtering out visible errors that do not affect the final outcome. 
\end{itemize}

\begin{figure*}[t]
    \centering
    \includegraphics[width=\textwidth]{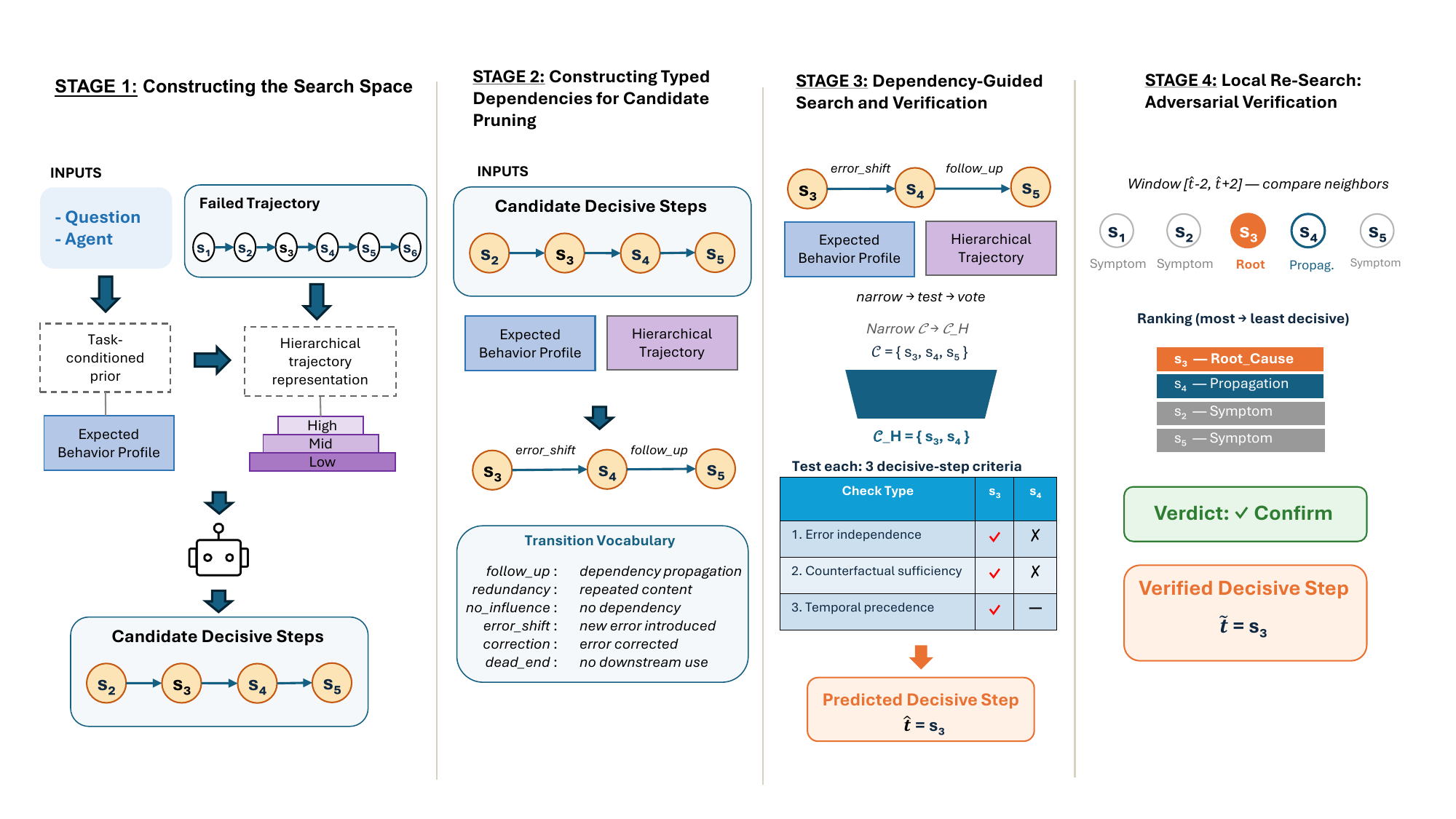}
    \caption{Overview of \tool. Stage 1 constructs an external prior $\pi$, a three-level trajectory representation $M$, and an initial candidate set $\mathcal{C}$. Stage 2 constructs typed dependencies to separate possible error sources from downstream carriers and prune candidates. Stage 3 performs dependency-guided search and verifies whether fixing a candidate would recover the expected output. Stage 4 locally verifies the predicted decisive step.}
    \label{fig:overall-approach}
\end{figure*}

\tool operationalizes these principles as a four-stage search procedure (Figure~\ref{fig:overall-approach}). The first stage \textbf{constructs the search space}: a task-conditioned prior $\pi$ and a hierarchical trajectory representation $M$ define what is searched and what counts as a deviation. The second stage \textbf{constructs typed dependencies for candidate pruning}: typed transitions determine which candidates can be pruned and which remain as possible error sources. The third stage \textbf{executes dependency-guided search}: the candidate set is narrowed and checked by asking whether fixing a candidate would recover the expected output. The fourth stage performs \textbf{local verification}: nearby candidates are re-examined to reduce premature commitment to a plausible but non-decisive step.

\paragraph{Notation.}
Let $\mathcal{G}$ denote an agentic system and $Q$ denote a task specification.
A trajectory $\tau=(s_1,s_2,\dots,s_T)$ is an ordered sequence of $T$ steps
generated by $\mathcal{G}$ in response to $Q$. Each step
$s_i=(r_i,u_i,o_i)$ consists of intermediate reasoning $r_i$, an action or
message $u_i$, and the resulting output or observation $o_i$. A trajectory is
classified as a failure if the final output $\hat{o}$ produced at $s_T$
differs from the expected output $o^*$, i.e., $\hat{o}\neq o^*$. A step $s_i$
is \emph{active} if it involves an independent decision, such as retrieval,
computation, filtering, tool use, or code execution, and \emph{passive} if it
only relays prior information. Let
$\mathcal{I}_{\mathrm{act}}(\tau)\subseteq\{1,\dots,T\}$ denote the indices of
active steps in $\tau$.

\paragraph{Definition (decisive step set).}
We formalize failure diagnosis as identifying a minimal subset of active steps
whose correction would be sufficient to recover the expected outcome. Formally,
a decisive step set $S^*$ satisfies
\[
    S^* \in \argmin_{S \subseteq \mathcal{I}_{\mathrm{act}}(\tau)} |S|
    \quad \text{s.t.} \quad
    \exists \{s_t'\}_{t \in S} :
    f\!\bigl(\tau^{(S \leftarrow S')}\bigr)=o^* .
\]
Here, $\tau^{(S \leftarrow S')}$ denotes the counterfactual trajectory obtained
by replacing each selected step $s_t$ with a corrected step $s_t'$, while
allowing subsequent steps to adapt to the corrected state. The function $f$
returns the final output induced by this counterfactual trajectory. When the
minimal set contains a single step, the decisive index is $t^*=\min S^*$,
which recovers the standard step-level formulation~\citep{zhang2025which}.
Throughout the paper, the search target is $S^*$ and the responsible agent(s)
associated with its steps, denoted by $\alpha^*$. For exposition, and to match
the benchmark annotation format, we focus on the common single-step case unless
otherwise noted.


\subsection{Constructing the Search Space}
\label{sec:Candidate-Step}

Effective searching in long trajectories requires a structured space in which coarse regions can be examined and pruned before fine-grained inspection. 


\paragraph{Defining an external prior.} 
%
%
We construct the external $\pi$ from the task specification $Q$, the agent set $\mathcal{A}$, the agents' roles and system prompts $\mathcal{P}$, the available tools $\mathcal{T}$, and the failed final output $\hat{o}$. 
The intermediate reasoning, actions, or tool outputs in $\tau$ are not used to ensure $\pi$ is independent of any intermediate steps. This independence is crucial because agentic failures can remain internally coherent: once an early error corrupts the state, later steps may be reasonable responses to that corrupted state. Evaluators based only on trajectory self-consistency may therefore miss the original deviation.  
The prior records the task objective, expected output type, appropriate agent/tool use at each stage, tool-use constraints, and task-specific failure risks such as unsupported information use, skipped verification, or incorrect intermediate values~\citep{bouzenia2025understanding}. 
These risks serve only as guidance for identifying suspicious steps, not as evidence of failure.


\paragraph{Hierarchical representation as a search space.}
Rather than searching directly over $\tau$, we construct a hierarchical abstraction that reduces the search space. Specifically, we convert $\tau$ into a three-level representation $M=(M_{\text{hi}},M_{\text{mid}},M_{\text{lo}})$ (Table~\ref{tab:hierarchical-representation}), conditioned on the external prior $\pi$. The high level captures coarse execution stages, the mid level summarizes local decisions and their significance, and the low level preserves raw execution evidence such as tool outputs and concrete values. This coarse-to-fine decomposition follows the principle of abstraction hierarchies in planning~\citep{sacerdoti1974planning,miki2015if} and temporal abstraction in reinforcement learning~\citep{sutton1999between}. It allows search to first identify suspect stages in $M_{\text{hi}}$, then localize candidate steps in $M_{\text{mid}}$, and finally verify them against evidence in $M_{\text{lo}}$.

\tool constructs a hierarchical representation $M$ from a trajectory $\tau = (s_1, \dots, s_T)$ via non-overlapping window-based abstraction. The trajectory is partitioned into $N = \lceil T/k \rceil$ consecutive windows $\{W_j\}_{j=1}^N$, where each $W_j$ contains $k$ steps, except possibly the final window, and is processed independently to ensure uniform coverage and reduce positional bias~\citep{liu2024lost, rafi2024order}.

For each window $W_j$, \tool retains the original steps as $M_{\text{lo}}^{(j)}=W_j$, summarizes them into mid-level representations $M_{\text{mid}}^{(j)}$, and further abstracts them into high-level representations $M_{\text{hi}}^{(j)}$. Both abstraction stages are conditioned on the prior $\pi$, so actions and decisions are interpreted relative to the expected task behavior rather than only their local text. The full representation is $M=\{(M_{\text{lo}}^{(j)},M_{\text{mid}}^{(j)},M_{\text{hi}}^{(j)})\}_{j=1}^N$. Here, $\pi$ guides abstraction, while the evidence stored in $M$ remains derived from the observed trajectory. The three levels are aligned, enabling top-down localization from coarse segments in $M_{\text{hi}}$ to raw evidence in $M_{\text{lo}}$. Processing windows independently trades limited cross-window context for substantially reduced positional bias in long trajectories.

\paragraph{Error candidate set selection.}
Using the prior $\pi$ and the hierarchical representation $M$, \tool identifies an initial error candidate set $\mathcal{C}$ of possible failure locations. The role of $M$ is to make candidate selection coarse-to-fine rather than purely step-wise: $M_{\text{hi}}$ identifies execution stages that deviate from the expected task progress, $M_{\text{mid}}$ identifies suspicious steps within those stages by summarizing each step's agent, action, decision, output, and role in the local context and $M_{\text{lo}}$ provides the raw evidence used to check whether the corresponding steps should enter the candidate set. Thus, a step $s_i$ is included in $\mathcal{C}$ if the aligned representations around $s_i$ show at least one signal of potential failure. At this stage, $\mathcal{C}$ is not treated as the decisive set; it only defines the initial search candidates for later typed-transition analysis and counterfactual verification.

Let $\tau = (s_1, \dots, s_T)$ and let $Y(\tau)$ denote the final outcome. For each step $s_i$, \tool evaluates three signals using the aligned views of $M$: (i) \emph{outcome relevance}, where $M_{\text{hi}}$ and $M_{\text{mid}}$ indicate that $s_i$ belongs to a stage or transition that contributes to $Y(\tau)$; (ii) \emph{prior inconsistency}, where the action, decision, or transition associated with $s_i$ conflicts with the task objective, agent role, tool-use constraints, or expected verification behavior encoded in $\pi$; and (iii) \emph{local execution inconsistency}, where $M_{\text{lo}}$ shows that the agent's stated reasoning, executed action, and observed output do not align, such as when an agent calls a tool inconsistent with its stated intent, uses mismatched tool arguments, ignores a tool error, or continues from an empty or unsupported result.

Let $\delta_i^{(r)}, \delta_i^{(p)}, \delta_i^{(e)} \in \{0,1\}$ denote
indicators for these conditions. The candidate set is then
\[
\mathcal{C} = \{ i \mid \delta_i^{(r)} \lor \delta_i^{(p)} \lor
\delta_i^{(e)} = 1 \}.
\]

Because $M$ is constructed over disjoint windows $\{W_j\}_{j=1}^N$, candidates are consolidated across windows by merging adjacent indices into contiguous regions $\mathcal{R} = \{R_k\}$, where each $R_k \subseteq \{1, \dots, T\}$ is a maximal interval of candidate steps. This consolidation uses the alignment between $M_{\text{hi}}$, $M_{\text{mid}}$, and $M_{\text{lo}}$ to recover context when suspicious behavior spans window boundaries.

\begin{table}
\centering
\small
\begin{tabularx}{\linewidth}{@{}lXX@{}}
\toprule
\textbf{Level} & \textbf{Contents} & \textbf{Role in Search} \\
\midrule
High ($M_{\text{hi}}$)
& Global execution summary: task objective, expected output type, high-level failure pattern, and agent responsibilities
& Identifies suspect regions of the trajectory and provides global context for dependency-guided pruning (\S\ref{sec:Candidate-Step}) \\

Mid ($M_{\text{mid}}$)
& Local step summaries: step index, agent, action, and decision, and significance in the trajectory
& Localizes suspicious steps within suspect regions and provides candidate-level context for constructing typed dependencies (\S\ref{sec:causal}) \\

Low ($M_{\text{lo}}$)
& Step-level evidence: original step content, tool outputs, concrete values, assumptions, and provenance
& Supports candidate verification against raw execution evidence (\S\ref{sec:diagnose})\\
\bottomrule
\end{tabularx}
\caption{Hierarchical trajectory representation as a three-level search space. Each level supports a distinct search operation, and the levels are aligned so that coarse-level narrowing constrains lower-level candidate selection and verification.}
\label{tab:hierarchical-representation}
\end{table}


\subsection{Constructing Typed Dependencies for Candidate Pruning}
\label{sec:causal}

The error candidate set $\mathcal{C}$ identifies steps where deviations may occur, but it does not distinguish between steps that \emph{originate} an error and those that merely \emph{propagate} it. This distinction is critical, as downstream steps often remain internally consistent with earlier errors and are thus incorrectly identified as decisive~\citep{zhang2025which}.

To separate error sources from downstream carriers and prune candidates whose effects cannot reach the final output, we construct a directed dependency structure over candidate steps. For each ordered pair $(s_i,s_j)$ with $i<j$, we assign a transition label $\ell_{i\rightarrow j}\in\mathcal{L}$, where $\mathcal{L}$ is a finite set of diagnostic relation types. These labels describe how information, decisions, and errors move between candidate steps. The resulting structure is used to retain candidates that may introduce or carry a deviation toward the final output $\hat{o}$, and to prune candidates whose effects are redundant, corrected, unrelated, or unable to reach the final output.

We use the following transition labels:
\begin{itemize}

\item \texttt{follow\_up}: $s_j$ builds on the output, decision, or assumption
introduced by $s_i$. If $s_i$ is aligned with the prior $\pi$, the transition
indicates normal progression and can be pruned. If $s_i$ is already corrupted,
$s_j$ is treated as a downstream carrier, and attribution is traced back to
$s_i$.

\item \texttt{redundancy}: $s_j$ repeats, restates, or rechecks information
already provided by $s_i$ without adding new evidence or changing the
trajectory. Such transitions are pruned unless the repeated step introduces an
independent deviation.

\item \texttt{no\_influence}: $s_j$ does not consume or depend on the output of
$s_i$. This relation prevents attribution from being transferred across
unrelated steps. Candidates connected only through such transitions are pruned
unless they independently deviate from the prior.

\item \texttt{error\_shift}: $s_j$ introduces a new deviation, shifts the
trajectory away from the expected path, or transforms an upstream issue into a
different downstream error. Such steps are retained as possible error sources
or error-shifting points.

\item \texttt{correction}: $s_j$ fixes, compensates for, or invalidates a
deviation introduced by $s_i$. Upstream candidates whose effects no longer
propagate beyond the correction are pruned.

\item \texttt{dead\_end}: $s_i$ leads to an output that is not consumed by later
reasoning or the final answer. Such candidates are pruned unless there is
independent evidence that they affect the final output.

\end{itemize}

Applying these labels yields a refined candidate set
$\mathcal{C}'\subseteq\mathcal{C}$ containing candidates whose effects can
plausibly reach the final output $\hat{o}$. This reduces the search space while
preserving candidates that may have originated, propagated, or transformed the
failure.

\subsection{Dependency-Guided Search and Verification}
\label{sec:diagnose}

After typed-transition structure, 
{\tool} further narrows $\mathcal{C}'$ to a small high-confidence set $\mathcal{C}_H$ and verifies each remaining candidate against the counterfactual sufficiency objective.

\paragraph{Narrowing the candidates.}
Ranking combines $\pi$ with the typed transition structure of $\mathcal{C}'$. Candidates are prioritized if they introduce a deviation from $\pi$ \emph{and} lie on a propagation path that reaches the final output. They are deprioritized if their outputs are unused, corrected, or independent of the final output (i.e., already pruned by \texttt{dead\_end}, \texttt{redundancy}, or \texttt{follow\_up} deductions). The remaining candidates are ranked using three signals: (i) whether the step makes an incorrect execution decision (e.g., wrong tool, flawed code, or misused data); (ii) whether it introduces or relies on unverified or fabricated information relative to $\pi$; and (iii) whether it lies on a typed transition path that influences subsequent reasoning or the final output. We retain the top $K$ candidates as the high-confidence set $\mathcal{C}_H$, with $K=3$ in our experiments. If fewer than $K$ candidates remain after pruning, all remaining candidates are included. 

\paragraph{Verification against the counterfactual objective.}
For each candidate in $\mathcal{C}_H$, \tool constructs a localized
verification context centered on the candidate: the prior $\pi$, the raw
window $(r,a,o)$ for steps $[i{-}1,\,i,\,i{+}1]$, the typed transitions
involving $i$, and linked $M_{\text{lo}}$ evidence. Where available, raw logs
such as executable code and full tool outputs replace summarized content to
preserve precise execution details. The full representation $M$ remains
available when broader context is required.

We then evaluate each candidate using three criteria aligned with the search
objective:
\begin{enumerate}
    \item \textbf{Error independence.} Does the step introduce the error, or
    merely inherit it from an upstream fault? We assess this by tracing typed
    dependencies backward from the candidate: \texttt{error\_shift} indicates
    that the candidate may introduce or transform an error, while
    \texttt{follow\_up} indicates propagation from an earlier step.

    \item \textbf{Counterfactual sufficiency.} Would correcting this step's
    output, decision, or assumption recover $o^*$ while keeping prior steps
    fixed and allowing subsequent steps to adapt?

    \item \textbf{Temporal precedence.} Among candidates satisfying the above
    criteria, select the earliest.
\end{enumerate}

The candidate that satisfies all three criteria is selected as the predicted
decisive step $\hat{t}$, and its corresponding agent is denoted by
$\hat{\alpha}$.


\subsection{Local Re-Search: Adversarial Verification}
\label{sec:verify}


A common issue in LLM-as-judge settings is \emph{anchoring}. Namely. once a plausible candidate is selected, later reasoning may rationalize it rather than challenge it. Thus, the predicted decisive step $\hat{t}$ may still be worse than a nearby alternative under direct comparison. We address this with \textbf{local verification}: a bounded re-evaluation that compares $\hat{t}$ against its immediate neighbors and re-applies the source--propagation distinction at finer scope.

Given $\hat{t}$, \tool extracts a bounded neighborhood $[\max(1,\hat{t}-2),\,\min(\hat{t}+2,T)]$ and provides the LLM with $\pi$, the relevant typed transitions, and $M$. For each step in the window, the LLM constructs arguments for and against the step being decisive and assigns one of four roles: \textsc{Root\_Cause}, which independently introduces the decisive error; \textsc{Propagation}, which carries an upstream error into downstream reasoning; \textsc{Symptom}, which reflects an existing error without materially affecting downstream computation; or \textsc{Contributing}, which affects execution but is not by itself sufficient to cause the final failure.

The four-role taxonomy is the local-search counterpart of the typed-transition vocabulary: \textsc{Root\_Cause} and \textsc{Propagation} are the cause-vs-carrier distinction restated as step-level roles, while \textsc{Symptom} and \textsc{Contributing} catch the two ways a step can look wrong without being decisive. \tool then ranks the steps in the window from most to least likely to be the decisive error using the role classification, the for/against arguments, the typed transition structure, and a counterfactual check on whether correcting the step would change the final output. This evaluation is comparative by construction since $\hat{t}$ must outperform specific named alternatives, which is precisely what local search requires to escape an anchored local optimum.

{\tool} then either \emph{confirms} that $\hat{t}$ remains the highest ranked decisive step or \emph{replaces} it with a higher ranked neighboring step. We denote the verified decisive step $\tilde{t}$ and its associated agent $\tilde{a}$.

\section{Experiment Setup}
\label{sec:exper_setup}
\noindent{\bf Benchmark.}
We evaluate \tool on Who\&When~\citep{zhang2025which}, a benchmark of 184 failed executions from 127 LLM-based multi-agent systems. The benchmark covers both CaptainAgent-generated teams on GAIA and AssistantBench tasks ~\citep{song2024adaptive,mialon2023gaia,yoran2024assistantbench} and hand-crafted Magentic-One systems~\citep{fourney2024magentic}. Each instance provides the full trajectory and human annotations for the responsible agent and decisive error step. 

\noindent{\bf Metrics.} Following prior work~\citep{zhang2025which}, we evaluate \tool using three metrics. \textbf{Agent-level accuracy} measures whether the predicted responsible agent $\hat{\alpha}$ matches the ground-truth agent $\alpha^*$. \textbf{Step-level accuracy} measures whether the predicted decisive step $\hat{t}$ exactly matches the ground-truth step $t^*$, making it the strictest metric. \textbf{Step-level accuracy with tolerance} considers a prediction correct if $|\hat{t}-t^*|\leq k$. We report results for $k\in\{1,2,3,4,5\}$.


\noindent{\bf Baselines.} We compare \tool with CHIEF~\citep{wang2026flat}, which we re-run under the same three backbone models as \tool, and AgentTracer-8B~\citep{zhang2025agentracer}, whose results are adopted from prior work. We also include direct-prompting LLM baselines from Who\&When, where the full trajectory is provided to the model in a single pass.

\section{Result}
\label{sec:result}

\begin{table}[t]
\centering
\caption{Performance on algorithm-generated and hand-crafted trajectories. We report agent-level accuracy (Agent) and exact step-level accuracy (Step). Best results within each block are bolded. For fair comparison, we re-run CHIEF under the same backbone configurations used by \tool.} 
\label{tab:algo_vs_handcrafted}
\small
\setlength{\tabcolsep}{4pt}
\begin{tabular}{l l cc cc}
\toprule
\multirow{2}{*}{\textbf{Backbone / Model}} 
& \multirow{2}{*}{\textbf{Method}}
& \multicolumn{2}{c}{\textbf{Algorithm-Generated} ($n=126$)}
& \multicolumn{2}{c}{\textbf{Hand-Crafted} ($n=58$)} \\
\cmidrule(lr){3-4} \cmidrule(lr){5-6}
& & \textbf{Agent} & \textbf{Step} & \textbf{Agent} & \textbf{Step} \\
\midrule

\multicolumn{6}{l}{\textit{Same-backbone comparison}} \\
GPT-5.4-Mini 
& CHIEF & 59.0 & 27.0 & 55.2 & 13.8 \\
& \tool & \textbf{66.5} & \textbf{42.0} & \textbf{65.5} & \textbf{22.2} \\

DeepSeek V3.2
& CHIEF & 52.0 & 30.0 & 50.0 & 10.3 \\
& \tool & \textbf{59.7} & \textbf{38.0} & \textbf{73.0} & \textbf{29.1} \\

MiniMax 2.5
& CHIEF & 54.8 & 37.3 & 51.0 & 17.0 \\
& \tool & \textbf{68.0} & \textbf{46.0} & \textbf{66.3} & \textbf{25.9} \\

\midrule
\multicolumn{6}{l}{\textit{Specialized attribution baseline} (results adopted from \cite{zhang2025agentracer})} \\
AgentTracer-8B  
& AgentTracer & 63.7 & 37.3 & 63.8 & 20.7 \\

\midrule
\multicolumn{6}{l}{\textit{Standalone LLM baselines} (results adopted from \cite{zhang2025agentracer})}\\
Qwen3-8B        & Direct & 60.3 & 5.6  & 39.5 & 3.5 \\
LLaMA-3.2-3B    & Direct & 45.2 & 8.7  & 50.0 & 3.5 \\
Qwen3-32B       & Direct & 57.9 & 8.7  & 44.8 & 1.7 \\
Qwen3-Coder     & Direct & 36.5 & 32.5 & \textbf{60.4} & 13.8 \\
GPT-4.1         & Direct & 59.5 & 21.9 & 37.9 & 3.4 \\
DeepSeek-R1     & Direct & \textbf{65.1} & 29.5 & 53.4 & 6.9 \\
Gemini-2.5-Pro  & Direct & 57.1 & 25.9 & 51.7 & 6.9 \\
Claude-Sonnet-4 & Direct & 51.1 & \textbf{38.8} & 50.0 & \textbf{18.9} \\
\bottomrule
\end{tabular}
\end{table}


\noindent{\bf Overall performance.} \tool achieves the strongest results on both trajectory types (Table~\ref{tab:algo_vs_handcrafted}). Its best configuration reaches 68.0 agent-level and 46.0 step-level accuracy on algorithm-generated trajectories, and 73.0 agent-level and 29.1 step-level accuracy on hand-crafted trajectories.
Compared with CHIEF and AgentTracer, \tool shows the largest gains at the step level, the stricter and more diagnostic metric. Against CHIEF, \tool improves agent-level accuracy by up to +13.2 points (MiniMax 2.5) on algorithm-generated trajectories and +23.0 points (DeepSeek V3.2) on hand-crafted trajectories. At the step level, \tool improves performance by +8.7 points (MiniMax 2.5) on algorithm-generated trajectories and +12.1 points (DeepSeek V3.2) on hand-crafted trajectories. Compared to AgentTracer, \tool further improves step-level accuracy by +8.7 points on algorithm-generated trajectories and +5.2 points on hand-crafted trajectories. These results indicate that \tool provides a more effective attribution procedure than existing trajectory-level baselines, especially for the harder step-localization task.


\noindent{\bf Comparison to standalone LLMs.}
Increasing model scale alone does not reliably solve step-level attribution. Small standalone models achieve near-zero step-level accuracy, while medium and large models remain unstable across trajectory types. The strongest standalone baseline, Claude-Sonnet-4, reaches 38.8 step-level accuracy on algorithm-generated trajectories but only 18.9 on hand-crafted trajectories. In contrast, \tool improves step-level accuracy by +7.2 points (38.8 $\rightarrow$ 46.0) on algorithm-generated trajectories and by +10.2 points (18.9 $\rightarrow$ 29.1) on hand-crafted trajectories. These results suggest that stronger backbones alone are insufficient for
reliable step-level attribution, and that structured search provides additional
benefits beyond model scaling 



\begin{table}[t]
\centering
\caption{Step-level accuracy under tolerance windows on Hand-Crafted trajectories. We report whether the predicted decisive step falls within $k$ steps of the ground-truth step ($|\hat{t}-t^*| \leq k$). }
\label{tab:trajfl_step_tolerance_handcrafted}
\small
\setlength{\tabcolsep}{4pt}
\begin{tabular}{l ccc}
\toprule
\multirow{2}{*}{\textbf{Tolerance}} 
& \multicolumn{3}{c}{\textbf{Hand-Crafted}} \\
\cmidrule(lr){2-4}
& \textbf{GPT-5.4-Mini} 
& \textbf{DeepSeek V3.2} 
& \textbf{MiniMax 2.5} \\
\midrule
Exact    & 22.2 & 29.1 & 25.9 \\
$\pm 1$ & 22.4 & 32.5 & 32.8 \\
$\pm 2$ & 25.9 & 36.7 & 34.5 \\
$\pm 3$ & 32.8 & 40.0 & 37.9 \\
$\pm 4$ & 41.4 & 45.6 & 51.7 \\
$\pm 5$ & 43.1 & 49.1 & 53.4 \\
\bottomrule
\end{tabular}
\end{table}

\noindent{\bf Performance under tolerance windows.} In practice, precisely identifying the decisive step is not always required; narrowing the failure to a range of steps is often sufficient. We therefore report tolerance-based step-level performance of \tool in Table~\ref{tab:trajfl_step_tolerance_handcrafted}, focusing on hand-crafted trajectories due to their higher difficulty. Algorithm-generated trajectories are excluded, as their short length (maximum of 10 steps) would lead to inflated accuracy under relaxed tolerances~\citep{zhang2025which}. As shown in Table~\ref{tab:trajfl_step_tolerance_handcrafted} exact step-level accuracy ($k=0$) remains modest (22.2--29.1), but increases steadily with relaxed tolerance, reaching up to 53.4 at $\pm 5$. This indicates that even when \tool does not predict the exact step, it typically localizes the error within a small neighborhood of the ground-truth decisive step.

\begin{table}[t]
\centering
\caption{Ablation results on algorithm-generated and hand-crafted trajectories.
All experiments use GPT-5.4-Mini as the backbone.} 
\label{tab:ablation_algo_handcrafted}
\small
\setlength{\tabcolsep}{4pt}
\renewcommand{\arraystretch}{1.08}
\begin{tabular}{lcccc}
\toprule
\multirow{2}{*}{\textbf{Configuration}} 
& \multicolumn{2}{c}{\textbf{Algorithm-Generated}} 
& \multicolumn{2}{c}{\textbf{Hand-Crafted}} \\
\cmidrule(lr){2-3} \cmidrule(lr){4-5}
& \textbf{Agent} & \textbf{Step} & \textbf{Agent} & \textbf{Step} \\
\midrule
Full \tool
& \textbf{66.5} & \textbf{42.0}
& \textbf{65.5} & \textbf{22.2} \\

\midrule
No external prior
& 55.6 {\scriptsize(-10.9)}
& 36.5 {\scriptsize(-5.5)}
& 58.6 {\scriptsize(-6.9)}
& 22.4 {\scriptsize(+0.2)} \\

No hierarchical representation
& 50.0 {\scriptsize(-16.5)}
& 31.7 {\scriptsize(-10.3)}
& 53.8 {\scriptsize(-11.7)}
& 17.2 {\scriptsize(-5.0)} \\

No transition typing
& 57.9 {\scriptsize(-8.6)}
& 37.3 {\scriptsize(-4.7)}
& 55.0 {\scriptsize(-10.5)}
& 18.8 {\scriptsize(-3.4)} \\
\bottomrule
\end{tabular}
\end{table}

\noindent{\bf Ablation study.} Table~\ref{tab:ablation_algo_handcrafted} shows how each component contributes to \tool's performance. The hierarchical representation has the largest impact. Removing it reduce step-level accuracy by 10.3 on algorithm-generated trajectories and 5.0 on hand-crafted trajectories. The external prior is also critical, particularly for algorithm-generated trajectories. Without it, agent-level and step-level accuracy decrease by 10.9 and 5.5, respectively. Removing transition typing yields consistent drops across both datasets, including a 10.5 decrease in agent-level accuracy on hand-crafted trajectories. Overall, these results suggest that \tool benefits from all three components, with the hierarchical representation playing the largest role.

\section{Limitations}
\label{sec:limitation}
\textbf{Dependence on LLM judgments.} \tool relies on LLM-based judgments at multiple stages, including candidate selection, transition typing, and counterfactual verification. As a result, its effectiveness depends on the underlying model’s reasoning reliability, and errors in intermediate judgments may propagate through the pipeline.

\textbf{Limited long-range dependency capture.} \tool constructs hierarchical abstractions and typed dependencies using window-based processing, which may miss long-range dependencies that span across windows. Although consolidation partially mitigates this issue, complex interactions across distant steps may not be fully captured.

\textbf{Assumption of localized failure.} Our formulation assumes that failures can be localized to a small set of decisive steps under a counterfactual correction objective. In scenarios where failures arise from distributed or interacting errors across multiple steps, this assumption may not fully hold, potentially limiting the precision of attribution.

\section{Related Work}
\label{sec:related_works}

\noindent{\bf LLM-based multi-agent systems.}
LLM-based multi-agent systems coordinate specialized agents through structured
communication, role assignment, and tool use~\citep{li2023camel,wu2023autogen,
hong2024metagpt,qian2024chatdev}. Recent work increasingly automates the
design of these systems, including prompt optimization~\citep{khattab2023dspy,
yuksekgonul2024textgrad}, routing and model selection~\citep{chen2025llmselector,
yue2025masrouter}, adaptive interaction topologies~\citep{zhuge2024gptswarm},
and workflow generation~\citep{li2025adaptive,zhang2024aflow,hu2024automated,
zhang2025multi,zhang2025evoflow}. These advances improve agent capability but
also produce long, interdependent trajectories in which errors can propagate
across agents, steps, and tool outputs. Our work focuses on diagnosing such
failures by reasoning over dependencies within a single failed trajectory.

\noindent{\bf Failure attribution and trajectory analysis.} Recent work has begun to study failures in multi-step LLM agent trajectories. Who\&When~\citep{zhang2025which} formalizes multi-agent failure attribution and shows that step-level localization remains challenging. Other studies analyze trajectory-level failure patterns such as loops, reasoning-action mismatches, and ineffective observation use~\citep{bouzenia2025understanding, banerjee2025did,he2025traject}, or introduce additional signals for debugging, including cross-run spectrum analysis~\citep{ge2025famas}, implicit execution traces~\citep{li2025implicit}, and graph-based interaction tracing ~\citep{zhang2025graphtracer}. Most related to our setting, AgenTracer ~\citep{zhang2025agentracer} learns an attribution model from synthetic counterfactual data, while CHIEF~\citep{wang2026flat} performs backtracking over a reconstructed causal graph. In contrast, \tool diagnoses a single failed trajectory without task-specific training data or oracle guidance. \tool explicitly types step-to-step dependencies to separate error origination from propagation and verifies candidates through counterfactual sufficiency.

\section{Conclusion}
\label{conclusion}

In this paper, we reformulated failure attribution in LLM agent trajectories as \emph{hierarchical dependency-guided search} rather than step-level error classification. Based on this formulation, we proposed {\tool}, a four-stage framework consisting of search space construction with a task-conditioned prior, candidate pruning through typed dependencies, dependency-guided search and verification, and local re-search. We evaluated {\tool} on the Who\&When benchmark, where it consistently outperformed existing SOTA baselines across multiple LLM backbones.

\bibliographystyle{unsrtnat}
\bibliography{references.bib}


\appendix
\label{appendix}



\section{Prompt Templates}
\label{app:prompts}
\definecolor{lightgraybox}{gray}{0.9}

\subsection{Prompts for Search-Space Construction}

\paragraph{External-prior prompt (\(\pi\) construction).}
This prompt operationalizes the construction of the external prior \(\pi\) from the task specification, agent roster, and observed incorrect output \(\hat{o}\), without exposing the intermediate trajectory. Its purpose is to define the expected output type, the reasoning requirements of the task, the role boundaries of each agent, and a task-specific set of failure risks that later guide candidate selection.

\begin{tcolorbox}[
    breakable,
    colback=lightgraybox,
    colframe=gray!60,
    boxrule=0.4pt,
    arc=2pt,
    left=6pt,
    right=6pt,
    top=6pt,
    bottom=6pt,
    fontupper=\footnotesize\ttfamily
]
\textbf{\# QUESTION}

\textbf{Question:} \texttt{\{question\}} \\
\textbf{Expected answer type:} \texttt{\{answer\_type\}} \\
\textbf{Constraints:} \texttt{\{constraints\}}

\vspace{0.5em}
\textbf{\# AGENTS}

\texttt{\{agents\_block\}}

\vspace{0.5em}
\textbf{\# AGENT'S FINAL ANSWER (WRONG)}

\texttt{\{final\_answer\}}

\vspace{0.5em}
\textbf{\# INSTRUCTIONS}

Produce the following sections. Stay observational; do not plan a solution.

\vspace{0.5em}
\texttt{===QUESTION\_ANALYSIS===}

- What is the question asking for? Be precise about the answer type.\\
- What kind of reasoning is required (e.g., retrieval, calculation, comparison, cross-referencing, verification)?\\
- What makes this question hard (e.g., ambiguity, multiple required data points, domain knowledge)?

\vspace{0.5em}
\texttt{===AGENT\_ROLES===}

For each agent, describe:\\
- What domain expertise they bring\\
- What they are equipped to do vs.\ what falls outside their stated role\\
- What a competent agent in this role should be expected to catch or verify

\vspace{0.5em}
\texttt{===ERROR\_SIGNALS===}

Based on the question type and the observed wrong final answer, what patterns should we watch for when reading the trajectory?

Examples of error signals:\\
- Data truncation (incomplete lists, pagination, ``showing \(N\) of \(M\)'')\\
- Cross-reference with incomplete data on one or both sides\\
- Fabricated data (values that appear without a retrieval or calculation source)\\
- Domain confusion (agent working outside its expertise)\\
- Verification theater (agent ``confirms'' by restating, not by checking)\\
- Wrong entity / wrong time period / off-by-one errors\\
- Placeholder values used as real data

\vspace{0.5em}
List \(3\)--\(5\) specific signals tailored to this question.
\end{tcolorbox}

\paragraph{Hierarchical-representation prompt (\(M\) construction).}
This prompt constructs the hierarchical representation \(M=(M_{\text{hi}}, M_{\text{mid}}, M_{\text{lo}})\) conditioned on the external prior \(\pi\). It organizes the observed trajectory into coarse execution summaries, local step summaries, and low-level evidence, which together define the initial search space.

\begin{tcolorbox}[
    breakable,
    colback=lightgraybox,
    colframe=gray!60,
    boxrule=0.4pt,
    arc=2pt,
    left=6pt,
    right=6pt,
    top=6pt,
    bottom=6pt,
    fontupper=\footnotesize\ttfamily
]
\textbf{\# CONTEXT ANALYSIS (your lens for reading this trajectory)}

\texttt{\{context\_analysis\}}

\vspace{0.5em}
\textbf{\# TASK CONTEXT}

\textbf{Question:} \texttt{\{question\}} \\
\textbf{Expected answer type:} \texttt{\{answer\_type\}} \\
\textbf{Constraints:} \texttt{\{constraints\}}

\vspace{0.5em}
\textbf{Agents involved:}

\texttt{\{agents\_block\}}

\vspace{0.5em}
\textbf{Agent's final answer:} \texttt{\{final\_answer\}}

\vspace{0.5em}
\textbf{Step-to-Agent index (\{num\_steps\} steps total):}

\texttt{\{step\_index\}}

\vspace{0.5em}
\textbf{\# FULL TRAJECTORY (\{num\_steps\} steps)}

\texttt{\{steps\_block\}}

\vspace{0.5em}
\textbf{\# INSTRUCTIONS}

Using the context analysis as your lens, read the trajectory and produce three memory documents. Focus on what is surprising, suspicious, or inconsistent, not on confirming expectations.

\vspace{0.5em}
\texttt{===HIGH\_LEVEL===}

Write \(4\)--\(6\) bullet points covering:\\
- What is the task asking?\\
- What did the agents answer?\\
- What overall strategy did the agents follow?\\
- What looks wrong at a glance?\\
- Which agent handled the most critical work?

\vspace{0.5em}
\texttt{===MID\_LEVEL===}

Write one entry per key moment. Skip all routine steps.

For each key moment:\\
\texttt{\#\#\# Step N (AgentName)}\\
What happened and why it matters. (\(2\)--\(3\) sentences max)

Include a step if:\\
- It computed or chose a value that feeds into the final answer\\
- It made a decision that changed the trajectory's direction\\
- Something looks wrong, suspicious, or surprising\\
- It introduced \textbf{NEW} information\\
- It matches an error signal from the context analysis

\vspace{0.5em}
\texttt{===LOW\_LEVEL===}

Write one entry per suspicious or critical step.

For each entry:\\
\texttt{\#\#\# Step N (AgentName)}\\
- \textbf{Exact values} computed or used\\
- \textbf{Data provenance}\\
- \textbf{Data completeness}\\
- \textbf{Assumptions made}\\
- \textbf{What specifically looks wrong or risky}\\
- \textbf{What inputs} this step received and whether they were correct
\end{tcolorbox}

\subsection{Prompts for Typed Dependency Construction}

\paragraph{Typed-dependency prompt.}
This prompt operationalizes typed dependency construction over diagnostically salient steps. It uses the hierarchical representation \(M\) to assign transition labels, detect loops and dead ends, and induce a value-flow structure that is later used to prune candidates from \(\mathcal{C}\) to \(\mathcal{C}'\).

\begin{tcolorbox}[
    breakable,
    colback=lightgraybox,
    colframe=gray!60,
    boxrule=0.4pt,
    arc=2pt,
    left=6pt,
    right=6pt,
    top=6pt,
    bottom=6pt,
    fontupper=\footnotesize\ttfamily
]
\textbf{\# HIERARCHICAL REPRESENTATION}

\textbf{\#\# High Level}

\texttt{\{high\_level\}}

\vspace{0.5em}
\textbf{\#\# Candidate Steps / Key Moments}

\texttt{\{mid\_level\}}

\vspace{0.5em}
\textbf{\#\# Supporting Evidence}

\texttt{\{low\_level\}}

\vspace{0.5em}
\textbf{\# STEP INDEX (all \{num\_steps\} steps, condensed)}

\texttt{\{step\_index\}}

\vspace{0.5em}
\textbf{\# INSTRUCTIONS}

Analyze how the candidate steps depend on one another. Your goal is to determine which candidates may have introduced, propagated, transformed, or eliminated a deviation that affects the final output.

For each ordered pair of candidate steps in temporal order:\\
1. Determine whether the later step depends on the earlier step's output, decision, or assumption.\\
2. Assign one transition label from the set below.\\
3. State briefly how the later step relates to the earlier step and whether this relation suggests source, propagation, correction, redundancy, or irrelevance.

\vspace{0.5em}
\textbf{\# OUTPUT FORMAT}

Write a single markdown document with the following sections:

\vspace{0.5em}
\textbf{\#\# Typed Dependencies}

For each relevant ordered pair:\\
\texttt{\#\#\# Step X \(\rightarrow\) Step Y}\\
- \textbf{Type}: \texttt{follow\_up | redundancy | no\_influence | error\_shift | correction | dead\_end}\\
- \textbf{Relationship}: 1 sentence on how \(Y\) depends on \(X\).

\vspace{0.5em}
\textbf{\# TRANSITION LABELS}

- \texttt{follow\_up}: step \(Y\) builds on the output, decision, or assumption introduced by step \(X\). If step \(X\) is already corrupted, treat step \(Y\) as a downstream carrier.\\
- \texttt{redundancy}: step \(Y\) repeats, restates, or rechecks information already provided by step \(X\) without adding new evidence or changing the trajectory.\\
- \texttt{no\_influence}: step \(Y\) does not consume or depend on the output of step \(X\).\\
- \texttt{error\_shift}: step \(Y\) introduces a new deviation, shifts the trajectory away from the expected path, or transforms an upstream issue at step \(X\) into a different downstream error.\\
- \texttt{correction}: step \(Y\) fixes, compensates for, or invalidates a deviation introduced by step \(X\).\\
- \texttt{dead\_end}: step \(X\) leads to an output that is not consumed by step \(Y\) or by the final answer.

\vspace{0.5em}
\textbf{\#\# Candidate-Pruning Summary}

Summarize which candidates remain plausible error sources or downstream carriers whose effects can still reach the final output, and which candidates can be pruned as redundant, corrected, unrelated, or dead ends.

\vspace{0.5em}
\textbf{\#\# Value Flow}

Describe how information and decisions propagate from retained candidates toward the final output. If the chain is incomplete or broken, state where and why.
\end{tcolorbox}

\subsection{Prompts for Dependency-Guided Search}

\paragraph{High-confidence candidate selection prompt (\(\mathcal{C}_H\)).}
This prompt narrows the refined candidate set \(\mathcal{C}'\) to a small high-confidence subset \(\mathcal{C}_H\) using the external prior, the typed dependency structure, and the compressed trajectory representation.

\begin{tcolorbox}[
    breakable,
    colback=lightgraybox,
    colframe=gray!60,
    boxrule=0.4pt,
    arc=2pt,
    left=6pt,
    right=6pt,
    top=6pt,
    bottom=6pt,
    fontupper=\footnotesize\ttfamily
]
\textbf{\# TASK CONTEXT}

\textbf{Question:} \texttt{\{question\}} \\
\textbf{Expected answer type:} \texttt{\{answer\_type\}} \\
\textbf{Final answer given:} \texttt{\{final\_answer\}}

\vspace{0.5em}
\textbf{Agents:}

\texttt{\{agents\_block\}}

\vspace{0.5em}
\textbf{Step-to-Agent mapping:}

\texttt{\{step\_index\}}

\vspace{0.5em}
\textbf{\# TRAJECTORY REPRESENTATION}

\textbf{\#\# Big Picture}

\texttt{\{high\_level\}}

\vspace{0.5em}
\textbf{\#\# Key Moments}

\texttt{\{mid\_level\}}

\vspace{0.5em}
\texttt{\{transitions\_section\}}

\vspace{0.5em}
\textbf{\# INSTRUCTIONS}

Identify the \(2\)--\(3\) steps most likely to contain the decisive error. Use the step-to-agent mapping to assign the correct agent to each step. Rank the candidates by likelihood, with the most suspicious step first.

\vspace{0.5em}
Output \textbf{only} the JSON object.
\end{tcolorbox}

\paragraph{Candidate verification prompt.}
This prompt verifies candidates in \(\mathcal{C}_H\) against the counterfactual sufficiency objective using localized raw evidence, typed dependencies, and linked low-level evidence.

\begin{tcolorbox}[
    breakable,
    colback=lightgraybox,
    colframe=gray!60,
    boxrule=0.4pt,
    arc=2pt,
    left=6pt,
    right=6pt,
    top=6pt,
    bottom=6pt,
    fontupper=\footnotesize\ttfamily
]
\textbf{\# TASK CONTEXT}

\textbf{Question:} \texttt{\{question\}} \\
\textbf{Final answer given:} \texttt{\{final\_answer\}}

\vspace{0.5em}
\textbf{Step-to-Agent mapping:}

\texttt{\{step\_index\}}

\vspace{0.5em}
\textbf{\# CANDIDATES TO VERIFY}

\texttt{\{candidates\_block\}}

\vspace{0.5em}
\textbf{\# LOCAL RAW EVIDENCE}

Only the candidate steps and their immediate neighbors are shown. Candidate steps are marked explicitly.

\texttt{\{raw\_steps\}}

\vspace{0.5em}
\textbf{\# LINKED LOW-LEVEL EVIDENCE}

\texttt{\{low\_level\}}

\vspace{0.5em}
\textbf{\# INSTRUCTIONS}

For each candidate, evaluate the following three criteria:

1. \textbf{Error independence.} Does this step introduce the error, or merely inherit it from an upstream fault? Use the typed dependency structure to trace whether the candidate is an error source or a downstream carrier.\\

2. \textbf{Counterfactual sufficiency.} If this step's output, decision, or assumption were corrected, would the final answer recover the expected output while keeping prior steps fixed and allowing subsequent steps to adapt?\\

3. \textbf{Temporal precedence.} Among candidates satisfying the above criteria, prefer the earliest step.

\vspace{0.5em}
Select the candidate that best satisfies these conditions and return the decisive error together with supporting reasoning.

\vspace{0.5em}
Output \textbf{only} the JSON object.
\end{tcolorbox}

\subsection{Prompts for Local Re-Search}

\paragraph{Local re-search prompt (adversarial verification).}
This prompt performs bounded local re-search around the predicted decisive step \(\hat{t}\), comparing it directly against neighboring candidates under an adversarial for-and-against evaluation.

\begin{tcolorbox}[
    breakable,
    colback=lightgraybox,
    colframe=gray!60,
    boxrule=0.4pt,
    arc=2pt,
    left=6pt,
    right=6pt,
    top=6pt,
    bottom=6pt,
    fontupper=\footnotesize\ttfamily
]
\textbf{\# TASK CONTEXT}

\textbf{Question:} \texttt{\{question\}} \\
\textbf{Final answer given:} \texttt{\{final\_answer\}} \\
\textbf{Agents:} \texttt{\{agents\_block\}} \\
\textbf{Total steps:} \texttt{\{total\_steps\}}

\vspace{0.5em}
\textbf{\# PRELIMINARY DIAGNOSIS}

Predicted decisive error: \texttt{Step \{\^t\}} by \texttt{\{\^a\}} --- \texttt{\{predicted\_desc\}}\\
Backward chain: \texttt{\{backward\_chain\}}

\vspace{0.5em}
\textbf{\# LOCAL WINDOW (\([\hat{t}-2,\,\hat{t}+2]\))}

\texttt{\{candidates\_block\}}

\vspace{0.5em}
\textbf{\# ROLE DEFINITIONS}

For each step \(X\) in the local window, assign one of the following roles:

- \texttt{ROOT\_CAUSE}: step \(X\) independently introduces the decisive error, such that correcting it would change the final output.\\
- \texttt{PROPAGATION}: step \(X\) consumes an upstream error and carries it into downstream reasoning or output.\\
- \texttt{SYMPTOM}: step \(X\) reflects an existing error without materially contributing to downstream computation.\\
- \texttt{CONTRIBUTING}: step \(X\) is independently problematic but is not by itself sufficient to change the final output.

\vspace{0.5em}
\textbf{\# INSTRUCTIONS}

\textbf{\#\# Step A: FOR-AND-AGAINST}

For each step in the local window:\\
1. give \(1\)--\(2\) arguments \textbf{for} it being the decisive error;\\
2. give \(1\)--\(2\) arguments \textbf{against} it being the decisive error;\\
3. assign a suspicion score in \([0,1]\);\\
4. assign one role: \texttt{ROOT\_CAUSE}, \texttt{SYMPTOM}, \texttt{PROPAGATION}, or \texttt{CONTRIBUTING}.

Be critical. Do not simply confirm the current diagnosis; challenge it against nearby alternatives.

\vspace{0.5em}
\textbf{\#\# Step B: COUNTERFACTUAL TEST}

For the top-\(3\) candidates by suspicion score:\\
- If only this step were corrected, would the final answer become correct?\\
- Provide one sentence of reasoning.

\vspace{0.5em}
\textbf{\#\# Step C: FINAL VERDICT}

Based on Steps A and B, select the step with the highest suspicion score that also passes the counterfactual test. Produce a ranked list of all local candidates and return either \texttt{CONFIRM} or \texttt{REVISE}.

\vspace{0.5em}
\textbf{\# OUTPUT FORMAT}

\texttt{\{}\\
\hspace*{1em}\texttt{"candidates": [}\\
\hspace*{2em}\texttt{\{}\\
\hspace*{3em}\texttt{"step": <int>,}\\
\hspace*{3em}\texttt{"agent": "<name>",}\\
\hspace*{3em}\texttt{"for": ["<point>", ...],}\\
\hspace*{3em}\texttt{"against": ["<point>", ...],}\\
\hspace*{3em}\texttt{"suspicion\_score": <float 0.0--1.0>,}\\
\hspace*{3em}\texttt{"classification": "<ROOT\_CAUSE | SYMPTOM | PROPAGATION | CONTRIBUTING>"}\\
\hspace*{2em}\texttt{\}}\\
\hspace*{1em}\texttt{],}\\
\hspace*{1em}\texttt{"counterfactual\_tests": [}\\
\hspace*{2em}\texttt{\{}\\
\hspace*{3em}\texttt{"step": <int>,}\\
\hspace*{3em}\texttt{"passes": <true | false>,}\\
\hspace*{3em}\texttt{"reasoning": "<1 sentence>"}\\
\hspace*{2em}\texttt{\}}\\
\hspace*{1em}\texttt{],}\\
\hspace*{1em}\texttt{"ranked\_steps": [<int>, <int>, ...],}\\
\hspace*{1em}\texttt{"final\_verdict": \{}\\
\hspace*{2em}\texttt{"decision": "<CONFIRM | REVISE>",}\\
\hspace*{2em}\texttt{"decisive\_step": <int>,}\\
\hspace*{2em}\texttt{"decisive\_agent": "<name>",}\\
\hspace*{2em}\texttt{"failure\_mode": "<wrong\_query | wrong\_logic | wrong\_interpretation | wrong\_assumption | wrong\_calculation | wrong\_filter | missed\_info | hallucination | other>",}\\
\hspace*{2em}\texttt{"description": "<1--2 sentences>",}\\
\hspace*{2em}\texttt{"confidence": "<low | moderate | high>",}\\
\hspace*{2em}\texttt{"reasoning": "<1--2 sentences on why you confirmed or revised>"}\\
\hspace*{1em}\texttt{\}}\\
\texttt{\}}

\vspace{0.5em}
Output \textbf{only} the JSON object.
\end{tcolorbox}

\section{Compute and Execution Setup}
\label{app:exec_setup}
All experiments are conducted using API-based inference. We use the OpenAI API and OpenRouter to access the underlying LLM backbones. No local model training or fine-tuning is required. The experiments can be reproduced by invoking the same APIs with the specified prompts, model configurations, and parameters described in this paper. Execution time depends on API latency and the number of LLM calls per trajectory, as \tool involves multiple stages (prior construction, hierarchical abstraction, dependency typing, and verification).

\end{document}